\documentclass[letterpaper]{article} % DO NOT CHANGE THIS
\usepackage{aaai24}  % DO NOT CHANGE THIS
\usepackage{times}  % DO NOT CHANGE THIS
\usepackage{helvet}  % DO NOT CHANGE THIS
\usepackage{courier}  % DO NOT CHANGE THIS
\usepackage[hyphens]{url}  % DO NOT CHANGE THIS
\usepackage{graphicx} % DO NOT CHANGE THIS
\usepackage{natbib}  % DO NOT CHANGE THIS AND DO NOT ADD ANY OPTIONS TO IT
\usepackage{caption} % DO NOT CHANGE THIS AND DO NOT ADD ANY OPTIONS TO IT
\usepackage{algorithm}
\usepackage{algorithmic}

\usepackage{booktabs}  
\usepackage{amssymb}
\usepackage{bm}
\usepackage{textcomp}
\usepackage{marvosym}
\usepackage{color}
\usepackage{multirow}
\usepackage{amsmath}
\usepackage{pdfpages}
\usepackage{newfloat}
\usepackage{listings}
\DeclareCaptionStyle{ruled}{labelfont=normalfont,labelsep=colon,strut=off} % DO NOT CHANGE THIS
\floatstyle{ruled}
\newfloat{listing}{tb}{lst}{}
\floatname{listing}{Listing}
\title{Improving Cross-modal Alignment with Synthetic Pairs \\ for Text-only Image Captioning}
\author{
    Zhiyue Liu\textsuperscript{\rm 1,\rm 2}\thanks{Corresponding author.}, Jinyuan Liu\textsuperscript{\rm 1}, Fanrong Ma\textsuperscript{\rm 1}
}
\affiliations{
    \textsuperscript{\rm 1}School of Computer, Electronics and Information, Guangxi University, Nanning, China\\
    \textsuperscript{\rm 2}Guangxi Key Laboratory of Multimedia Communications and Network Technology\\
    liuzhy@gxu.edu.cn, \{2213394017, 2213301037\}@st.gxu.edu.cn
}

\usepackage{bibentry}
\begin{document}

\maketitle

\begin{abstract}
Although image captioning models have made significant advancements in recent years, the majority of them heavily depend on high-quality datasets containing paired images and texts which are costly to acquire. Previous works leverage the CLIP's cross-modal association ability for image captioning, relying solely on textual information under unsupervised settings. However, not only does a modality gap exist between CLIP text and image features, but a discrepancy also arises between training and inference due to the unavailability of real-world images, which hinders the cross-modal alignment in text-only captioning. This paper proposes a novel method to address these issues by incorporating synthetic image-text pairs. A pre-trained text-to-image model is deployed to obtain images that correspond to textual data, and the pseudo features of generated images are optimized toward the real ones in the CLIP embedding space. Furthermore, textual information is gathered to represent image features, resulting in the image features with various semantics and the bridged modality gap. To unify training and inference, synthetic image features would serve as the training prefix for the language decoder, while real images are used for inference. Additionally, salient objects in images are detected as assistance to enhance the learning of modality alignment. Experimental results demonstrate that our method obtains the state-of-the-art performance on benchmark datasets.

\end{abstract}

\section{Introduction}
Image captioning is an appealing research task, aiming to produce descriptive textual content for provided images. The major challenge in this task involves comprehending the association between images and texts, which heavily relies on paired datasets~\cite{stefanini2022show}. Current researches primarily concentrate on training models using paired data, leading to substantial enhancements~\cite{wu2022difnet,luo2023semantic,kuo2023haav}, while these models generalize poorly to images in the wild~\cite{wu2018decoupled,agrawal2019nocaps}. Despite pre-training on web-scale noisy paired data and subsequent fine-tuning on human-annotated captioning data have shown effectiveness~\cite{radford2021learning,51452,alayrac2022flamingo}, pre-trained methods still count on paired data to achieve optimal performance. The generalization ability is crucial in practical applications, especially in the absence of in-domain image-text paired samples. Unsupervised transferability enables the model to work on new tasks or domains without paired training data, making it a valuable feature for various real-world scenarios.

Unsupervised image captioning, as another line of work, has garnered increasing attention for its ability to generate captions without human-annotated data. 
By capitalizing on the multimodal embedding space of CLIP~\cite{radford2021learning}, recent methods adopt text-only training paradigms to develop a language model using CLIP text features as the input. During inference, they decode captions from the given image's features extracted by CLIP. In contrast to other unsupervised methods, text-only image captioning utilizes textual data during training without the accessibility to real-world images~\cite{li2023decap}. Although previous methods have shown competitive results, they heavily depend on the CLIP's cross-domain capability. The presence of the modality gap~\cite{liang2022mind,jiang2023understanding} poses a challenge in those text-only methods based on CLIP, as it hinders the direct use of visual embeddings as the source to generate captions. Towards this issue, some works propose solutions by noise injection or feature projection~\cite{nukrai-etal-2022-text,li2023decap}, assisting image features from CLIP to better align with text features during inference, while the modality gap still exists. Due to the absence of images, only textual information could be utilized for training while real images for inference. This leads to a discrepancy between training and inference. Thus, the learned alignment cannot work efficiently for real image captioning, hindering existing models from performing well. Besides, overreliance on CLIP would indeed obstruct the acquisition of valuable knowledge from other pre-trained models.

This paper proposes a method by exploring synthetic pairs for text-only image captioning, called SynTIC. With multiple frozen pre-trained models, we mitigate the inconsistency of training and inference, while better aligning texts and images. Firstly, a pre-trained text-to-image model is employed to generate the synthetic image corresponding to each training text. Nonetheless, directly training the captioning model on synthetic image-text pairs results in limited performance due to the dissimilarity between synthetic images and natural images. Specifically, images generated from simple descriptive texts are usually short of the depth and details in real-world images, indicating these generated images lack content richness. The cross-modal alignment acquired from synthetic data may compromise the model's ability to capture the semantics of real images. Therefore, we employ a generation-then-refinement procedure to synthesize image features. The feature of generated images from CLIP acts as initialization. Then, these features are optimized in the CLIP multimodal embedding space. Since CLIP is pre-trained on real-world pairs with the contrastive constraint, the optimization following this constraint helps synthetic image features embed in the CLIP multimodal space~\cite{radford2021learning}. This would polish pseudo image features to relatively resemble natural image features. We further refine pseudo features with various textual information. A text supporting set is deployed to project synthetic images as text embeddings. By leveraging the projected vectors, the caption decoder effectively assimilates information from the text corpus. To some extent, the limited semantics of synthetic image features are complemented by textual data, enabling our method to create diverse and contextually grounded descriptions for given images. 
Based on the above steps, the training and inference of SynTIC would follow a unified procedure that takes images as input and then projects image features into a text space for decoding. Moreover, our method introduces object tags detected in images as assistance to ease the learning of semantic alignments between modalities, since projected vectors in the text space may lack parts of semantics in the origin images. Those objects potentially contain essential information that serves as caption contents~\cite{li2020oscar}. In our method, object features would cooperate with image features to guide the decoder for captioning. SynTIC entirely relies on texts and corresponding synthetic images for training, without accessing any real images.

In summary, our contributions are as follows:
\begin{itemize}
\item A text-only captioning method, called SynTIC, is proposed, which leverages synthetic data to improve modality alignment, while unifying training and inference.

\item A generation-then-refinement procedure, which includes image feature optimization and projection, is developed to align images with texts. Besides, object tags serve as auxiliary points to enhance captioning performance.

\item Experimental evaluations on several benchmark datasets demonstrate that SynTIC enhances text-only training and achieves competitive generation performance.

\end{itemize}

\section{Related Work}
Depending on human-annotated image-text data, advanced methods for supervised captioning mainly work on visual encoding, language modeling, and training algorithms to boost captioning performance~\cite{gao2019self,xing2021km,li2022er,zeng2022s2}. Due to the lack of supervised datasets in specific scenes, some works pre-train models on the large-scale web-crawled dataset and then fine-tune on a few downstream human-annotated pairs~\cite{hu2022scaling,pmlr-v162-li22n}. With the rise of pre-trained multimodal methods (e.g., CLIP), they are deployed to improve captioning performance. Such as~\citeauthor{dai2022enabling}~(\citeyear{dai2022enabling}) attempt to directly migrate the knowledge from CLIP to image captioning.~\citeauthor{wang2022simvlm}~(\citeyear{wang2022simvlm}) realize text generation by the autoregression pre-training on image and text data. Although these methods get improvements from the pre-training stage, their optimal performance still relies on the fine-tuning stage which requires human-annotated data.

Under the unsupervised setting, the model aims to generate image captions without paired data. Unpaired captioning methods use independent image sources and text corpus for training. They mostly conduct adversarial training or detect objects to establish the connection between images and texts~\cite{ben2021unpaired,meng2022object,10097833,10.1145/3581783.3611891}. Without training, ZeroCap~\cite{tewel2022zerocap} proposes to integrate CLIP and a language model, where CLIP assumes the role of guiding the language model toward a specific visual direction. MAGIC~\cite{su2022language} develops a framework for incorporating visual controls into the generation process, enabling language models to perform zero-shot captioning. With no need for real-world images, text-only captioning methods have emerged to train a decoder solely on textual data, achieving superior unsupervised performance. They could generate captions based on feature alignment in the CLIP embedding space. Considering that the modality gap~\cite{liang2022mind} existing between CLIP text and image features, some works~\cite{nukrai-etal-2022-text, gu2023can} propose the solution by injecting noise to train a model that maps the text embedding closer to its corresponding image embedding. Besides, \citeauthor{li2023decap}~(\citeyear{li2023decap}) introduce a training-free mechanism that leverages a text memory set to project visual embeddings into the text embedding space during inference. Nevertheless, previous methods cannot completely narrow the modality gap. Using texts for training while images for inference still brings a discrepancy, which impedes effective text-image alignment. Moreover, the excessive dependence on the cross-modal capability of CLIP would hinder the acquisition of valuable knowledge from other pre-trained models. In contrast, our method synthesizes image features to unify training and inference following a generation-then-refinement procedure. CLIP would cooperate with the pre-trained text-to-image model in SynTIC, where synthetic features would be refined in the CLIP embedding space for better cross-modal alignment. Objects detected in images help further improve performance.

\begin{figure*}[t]
\centering
\includegraphics[width=1.95\columnwidth]{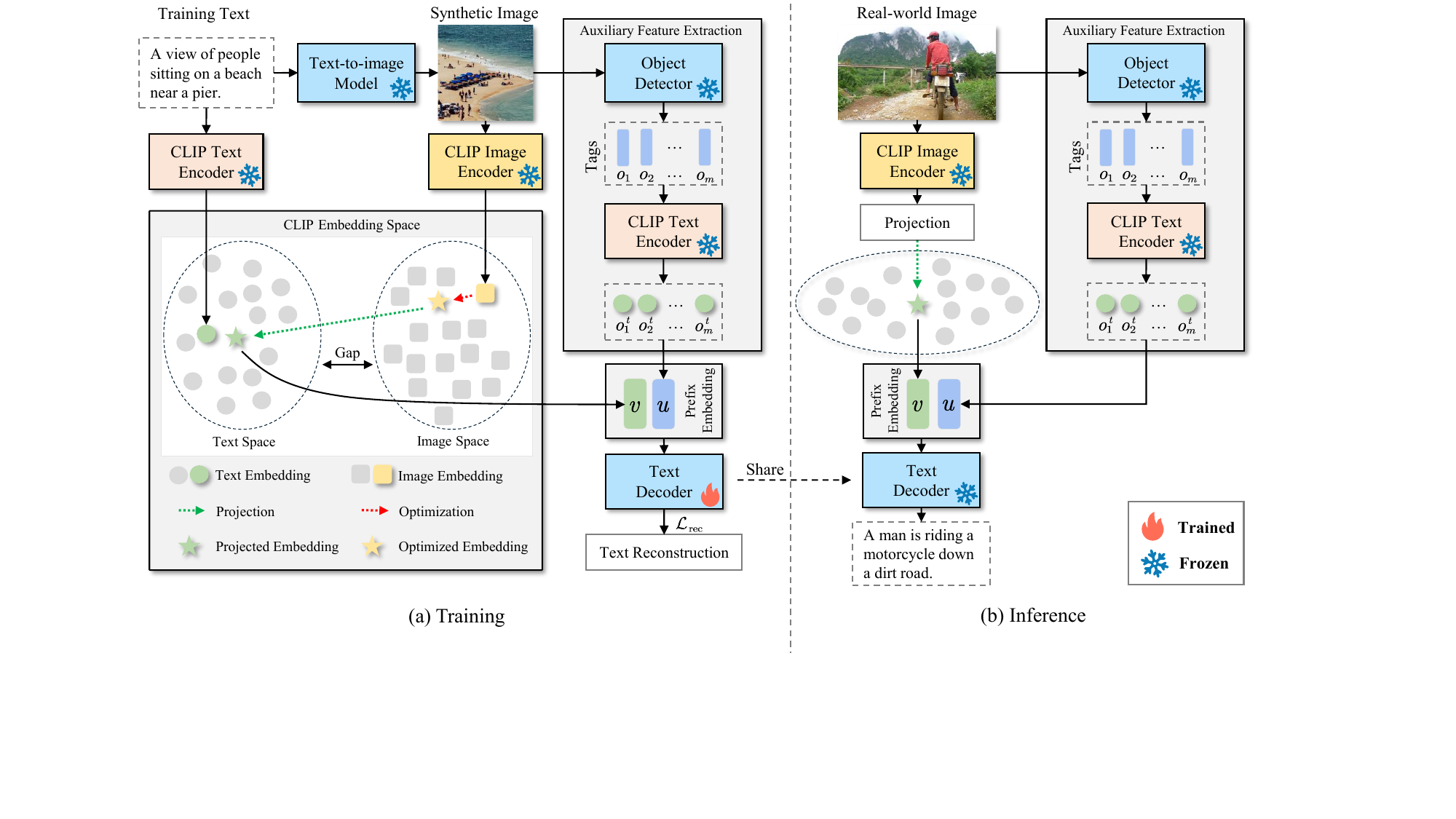}
    \caption{Overview of our proposed method. (a) shows the training procedure. A generated image corresponding to the training text is encoded and then refined in the CLIP embedding space, leading to improved cross-modal alignment. Objects detected in the image guide a caption decoder for text reconstruction. Except for the decoder, other pre-trained models are frozen. (b) shows the inference procedure. A real-world image is encoded and projected in the same way as described in (a), mitigating the discrepancy between training and inference. Then, objects from this natural image help the decoder obtain a caption.}
    \label{fig1}
\end{figure*}

\section{Methodology}
Text-only captioning requires training the model to generate captions for given images when textual input is solely available. Unlike fully supervised methods trained on the dataset $D_{\rm pair}=\{(i_1,t_1),...,(i_z,t_z)\}$ including pairs between the image $i_i$ and text $t_i$, text-only training could only use a text corpus $D_{t}=\{t_1,...,t_n\}$ as the source. The trained model would perform caption generation on the real-world images $D_{v}=\{i_1,...,i_m\}$ during inference. Achieving cross-modal alignment from only textual data poses a major challenge.

\subsection{Overview}
The overall framework of SynTIC is shown in Fig.~\ref{fig1}. Our method aims to improve cross-modal alignment by introducing synthetic image-text pairs. We refer to the embeddings encoded by CLIP as text and image features. During training, a pre-trained text-to-image model is utilized to generate related images for the texts within the corpus. Then, pseudo features of these images are optimized in the CLIP embedding space to enhance the alignment between synthetic images and their corresponding texts. We extract the features of all texts to construct a support set, and project the optimized pseudo features into the text embedding space. The objects occurring in images render an anchor point that strengthens cross-modal alignment. A lightweight structure serves as the language decoder, taking both projected features and auxiliary objects as inputs. For inference, real images are directly encoded and projected as the decoder's input.

\subsection{Pseudo Image Feature Synthesis}
The modality gap in the CLIP embedding space prevents the language decoder from directly utilizing image features as input because of its training on text features. The lack of images under text-only captioning causes the discrepancy between training and inference. Toward this issue, we exploit a generation-then-refinement method, which firstly constructs initial pseudo features from synthetic images, and then refine those features for better image-text alignment.

\subsubsection{Synthetic Image Generation.}
Given a text corpus $D_{t}=\{t_1,...,t_n\}$ with $n$ texts, we leverage the generative capacity of a pre-trained text-to-image model $G$ to obtain a synthetic image $s_i=G(t_i)$ for each $t_i$. But directly training the image captioning model using synthetic image-text pairs results in limited performance. This arises due to the fact that real images are depicted with multi-faceted details, while synthetic images generated from simple text tend to lack diverse contexts. Considering the satisfactory quality of those synthetic images, we adopt their CLIP features as initialization.

\subsubsection{Feature Optimization.}
Since CLIP is pre-trained on web-crawled data, its output features indicate the association of real-world texts and images. Optimizing the pseudo image features in the CLIP multimodal space could lead to better alignment with their paired texts, which may narrow the disparity between the pseudo features and real features. Specifically, the pre-trained CLIP encoder outputs a pseudo image feature $s^v_i = {\rm CLIP}_v(s_i)$ for the synthetic image $s_i$ and a text feature $t^t_j = {\rm CLIP}_t(t_j)$ for the text $t_j$, where ${\rm CLIP}_v$ and ${\rm CLIP}_t$ denote the image encoder and text encoder, respectively. The similarity between $s^v_i$ and $t^t_j$ is calculated as:
\begin{equation}
    f_{\rm sim}(s^v_i, t^t_j)=\frac{s^v_i}{||s^v_i||}\cdot\frac{t^t_j}{||t^t_j||}.
\end{equation}
Afterward, the pseudo image feature $s^v_i$ would be optimized by a contrastive loss as follows:
\begin{equation}
    {\nabla}_{s_i^v}\mathcal{L}_{\rm con}={\nabla}_{s_i^v}[-\frac{1}{b}\sum_{i=1}^{b} {\rm log}\frac{{\rm exp}({f_{\rm sim}(s^v_i, t^t_i)/\tau})}{\sum_{j=1}^b{\rm exp}({f_{\rm sim}(s^v_i, t^t_j)/\tau})}],
    \label{eq2}
\end{equation}
where $b$ is the mini-batch size and $\tau$ is a temperature parameter. Intuitively, Eq.~\ref{eq2} enforces that the pseudo image feature $s^v_i$ exhibits high cosine similarities with its corresponding text features $t^t_i$ within the multimodal feature space of CLIP, while having low similarities with other features. That is, refining $s^v_i$ through the contrastive loss essentially distills the knowledge of real-world image-text correspondence.

\subsubsection{Feature Projection.}
To further enrich the semantics of the optimized pseudo image feature and reduce the modality gap in the CLIP embedding space, SynTIC assembles abundant textual semantics and represents the pseudo image feature in the text embedding space. Since the contrastive loss shapes the cross-modal association, the CLIP text features could be employed to grab image representations~\cite{zhou2022lafite2}. Specifically, we reserve the text feature $t^t_k = {\rm CLIP}_t(t_k)$ for each $t^t_k$ in the corpus $D_t$ to construct the projection support set $D^s_t=\{t^t_1,...t^t_n\}$. The pseudo image feature is projected by a weighted combination of all features in this support set. SynTIC utilizes the cross-modal capability of CLIP and expresses the weight depending on the cosine similarity $f_{\rm sim}$ of two modalities. The projected vector of the pseudo image feature $s^v_i$ could be obtained as follows:
\begin{equation}
\resizebox{1\columnwidth}{!}{$
\begin{aligned}
           v_i=f_{\rm proj}(s_i^{v},D_t^s) & =\sum_{j=1}^n{w_j\ast t_j^{t}}\\
           & =\sum_{j=1}^n{\frac {{\rm exp}({f_{\rm sim}(s^v_i, t^t_j)/\tau})}{\sum_{k=1}^n{{\rm exp}({f_{\rm sim}(s^v_i, t^t_k)/\tau})}}}{\ast}t^t_j,
\end{aligned}
$}
    \label{eq3}
\end{equation}
where $w_j$ denotes the weight of $j$-th text feature $t^t_j$. Since $v_i$ is a combination of multiple text features, it contains abundant textual semantics. We utilize this projected vector in the text embedding space as the image feature which serves as a condition to decode a caption.

\subsection{Auxiliary Feature Extraction}
Salient objects in images could be identified by pre-trained object detectors, and these detected objects are frequently referenced in the paired text. In contrast to the visual object features from over-sampled regions, our method introduces object tags detected in images as auxiliary features to facilitate the process of learning semantic alignments between images and captions. Specifically, textual object information is regarded as a clue for caption decoding. Given a synthetic image $s_i$, we employ an object detection model to extract objects $O_i=\{o_1,...,o_m\}$ in the image. Then, the CLIP text encoder could obtain a text feature $o^t_k$ for all objects as:
\begin{equation}
    O^t_i=\{o^t_1,...,o^t_m\}=\{{\rm CLIP}_t(o_1),...,{\rm CLIP}_t(o_m)\},
\end{equation}
where $m$ is the number of objects detected from the image. Since the pseudo image feature would change after our contrastive optimization, we integrate $\{o^t_1,...,o^t_m\}$ by the multi-head attention mechanism~\cite{vaswani2017attention}. The auxiliary feature $u_i$ for $s_i$ is obtained as follows:
\begin{equation}
u_i={\rm MHAttn}(s_i^v,O_i^t,O_i^t),
\end{equation}
where ${\rm MHAttn}(Q,K,V)$ denotes the multi-head attention for query $s_i^v$, key $O_i^t$, and value $O_i^t$. The presence of textual information related to the objects in $u_i$ serves as an anchor, that assists cross-modal alignment.

\subsection{Prefix Decoding}
The decoder architecture, relying on the Transformer models, has found widespread application in various natural language generation tasks. Thus, following previous works~\cite{li2023decap}, our method employs a lightweight Transformer-based decoder as the generation model. Both image features and auxiliary object features should be involved in decoding. Considering these features as the prefix embedding, we train the decoder with prefix language modeling. The model would predict the next token conditioned on input features in an autoregressive manner.

\subsubsection{Training.} 
Given a caption text $t=\{w_1,...,w_{|t|}\}$, we obtain the corresponding projected image feature $v$ and auxiliary object feature $u$ as the prefix embedding. Then, the decoder is trained to reconstruct the caption text $t$ conditioned on the prefix, which minimizes the objective as follows:
\begin{equation}
    \mathcal{L}_{\rm rec}=-\frac{1}{|t|}\sum_{i=1}^{|t|} {\rm log}P_{\theta}(w_i|w_{<i},v,u),
\end{equation}
where $|t|$ denotes the sequence length of $t$, and $\theta$ is the trainable parameter of the decoder. The caption decoder is trainable, while implemented pre-trained models are frozen.
\subsubsection{Inference.}
Our method could unify training and inference by utilizing synthetic paired data. For a provided real-world image, we project it as the image feature using the textual support set, and simultaneously pick up the image's object tags that serve as the auxiliary feature. These image and object features are prefixes for the trained decoder to generate caption tokens step by step.

\begin{table*}[t]
\centering
\resizebox{1.6\columnwidth}{!}{
\begin{tabular}{l|cc|ccccc|ccccc}
\toprule
\multirow{2.5}{*}{Method}&\multicolumn{2}{c|}{Data} &\multicolumn{5}{c|}{MSCOCO}&\multicolumn{5}{c}{Flickr30K}\\
\cmidrule{2-13}
  & \textbf{I}  & \textbf{T}  & \textbf{B@4}($\uparrow$) & \textbf{M}($\uparrow$) & \textbf{R-L }($\uparrow$) & \textbf{C}($\uparrow$) & \textbf{S}($\uparrow$) &\textbf{B@4}($\uparrow$) & \textbf{M}($\uparrow$) & \textbf{R-L }($\uparrow$) & \textbf{C}($\uparrow$) & \textbf{S}($\uparrow$) \\
\midrule
\multicolumn{13}{c}{Supervised Approach}\\
\midrule
BUTD  &\checkmark &\checkmark & 36.2  & 27.0 &56.4&  113.5 & 20.3  & 27.3 &21.7&- & 56.6 &16.0 \\
ClipCap &\checkmark &\checkmark  & 33.5  & 27.5 &-&  113.1 & 21.1 & -  & - &-&- &- \\
Barraco et al. &\checkmark &\checkmark  &36.0  & 27.8 &56.5&  114.9 & 20.8 & -  & - &-&- &- \\
\midrule
\multicolumn{13}{c}{Unsupervised Approach}\\
\midrule
Feng et al. &\checkmark &\checkmark  & 18.6  &17.9 &43.1 &54.9   &11.1 &-  &- &-&- &- \\
Laina et al. &\checkmark &\checkmark  & 19.3  & 20.2 &45.0 &61.8 &12.9 &-  &- &-&- &- \\
ESPER-Style  &\checkmark &\checkmark  & 21.9  &21.9 &- &78.2 &- &-  &- &-&- &- \\
CLIPRe & &\checkmark  & 4.9  & 11.4 &29.0&  13.6 & 5.3 & 5.2  & 11.6 &27.6&  10.0 & 5.7 \\
ZeroCap & &\checkmark   & 7.0  & 15.4 &31.8&  34.5 & 9.2 & 5.4  & 11.8 &27.3&  16.8 & 6.2 \\
MAGIC  & &\checkmark & 12.9  & 17.4 &39.9&  49.3 & 11.3 & 6.4  & 13.1 &31.6&  20.4 & 7.1 \\
CapDec~~~ & &\checkmark & 26.4  & 25.1& 51.8&  91.8 &11.9 &17.7 &20.0 &43.9 &39.1 &9.9 \\
CLOSE & &\checkmark &28.6 &25.2 &- &95.4 &18.1 &- &- &- &- &- \\
DeCap & &\checkmark & 24.7  & 25.0 & - &  91.2 &18.7  &21.2 &21.8 &- &\textbf{56.7} &15.2 \\
\midrule
SynTIC (Ours) & &\checkmark & \textbf{29.9}  & \textbf{25.8} & \textbf{53.2} &  \textbf{101.1} &\textbf{19.3}  &\textbf{22.3} &\textbf{22.4} &\textbf{47.3} &56.6 &\textbf{16.6}\\
\bottomrule
\end{tabular}
}
\caption{In-domain image captioning results on MSCOCO and Flickr30K, where \textbf{B@4}, \textbf{M}, \textbf{R-L}, \textbf{C}, and \textbf{S} represent BLEU@4, METEOR, ROUGE-L, CIDEr, and SPICE, respectively. \textbf{I} and \textbf{T} denote training with image data and text data, respectively. $\uparrow$ means higher is better. The best results are in bold.}
\label{tab1}
\end{table*}

\begin{table*}[t]
\centering
\resizebox{1.4\columnwidth}{!}{
\begin{tabular}{l|ccccc|ccccc}
\toprule
\multirow{2.5}{*}{Method} &\multicolumn{5}{c|}{MSCOCO$\rightarrow$Flickr30K}&\multicolumn{5}{c}{Flickr30K$\rightarrow$MSCOCO}\\
\cmidrule{2-11}
  & \textbf{B@4}($\uparrow$) & \textbf{M}($\uparrow$) & \textbf{R-L }($\uparrow$) & \textbf{C}($\uparrow$) & \textbf{S}($\uparrow$) &\textbf{B@4}($\uparrow$) & \textbf{M}($\uparrow$) & \textbf{R-L }($\uparrow$) & \textbf{C}($\uparrow$) & \textbf{S}($\uparrow$) \\
\midrule
CLIPRe    & 4.4  & 9.6 &27.2&  5.9 & 4.8 & 3.0  & 9.9 &22.8& 8.5 & 3.9 \\
MAGIC   & 6.2 &12.2 &31.3 &17.5 &5.9 & 5.2 &12.5 &30.7 &18.3 &5.7 \\
CapDec~~~ & 17.3 &18.6 &42.7 &35.7 &7.2 &9.2 &16.3 &36.7 &27.3 &10.4 \\
DeCap & 16.3 &17.9&- &35.7 &11.1  &12.1 &18.0 &- &44.4 &10.9 \\
DeCap-TT & 17.7 & \textbf{20.2}&- &42.0 &13.8  &19.7 &20.9 &- &63.1 &13.9 \\
\midrule
SynTIC &{17.9}  &{18.6} &{42.7} &{38.4} &{11.9}  & {14.6}  & 19.4 & 40.9 &  {47.0} &{11.9}\\
SynTIC-TT & \textbf{19.4}  & \textbf{20.2} & \textbf{44.8} &  \textbf{43.2} &\textbf{13.9}  & \textbf{20.6}  & \textbf{21.3} & \textbf{46.6} &  \textbf{64.4} &\textbf{14.3}\\
\bottomrule
\end{tabular}
}
\caption{Cross-domain image captioning results, where X$\rightarrow$Y means source domain $\rightarrow$ target domain. }
\label{tab2}
\end{table*}

\begin{table}[t]
\centering
\resizebox{0.9\columnwidth}{!}{
\begin{tabular}{l|c|ccccc}
\toprule
\multirow{2.5}{*}{Method}&\multirow{2.5}{*}{Dataset} &\multicolumn{5}{c}{MSCOCO}\\
\cmidrule{3-7}
  & & \textbf{B@4}($\uparrow$) & \textbf{M}($\uparrow$) & \textbf{R-L }($\uparrow$) & \textbf{C}($\uparrow$) & \textbf{S}($\uparrow$)\\
\midrule
ZeroCap &-   & 2.6  & 11.5 &-&  14.6 & 5.5\\
CLIPRe  &CC3M & 4.6 &13.3 &- &25.6 &9.2\\
DeCap &CC3M & 8.8 &16.0 &- &42.1 &10.9\\
DeCap &SS1M &8.9 &17.5 &- &50.6 &13.1\\
\midrule
SynTIC &SS1M & \textbf{13.3}  & \textbf{17.6} & \textbf{36.7} &  \textbf{55.6} &\textbf{13.3} \\
\bottomrule
\end{tabular}
}
\caption{Zero-shot image captioning results on MSCOCO. }
\label{tab3}
\end{table}

\section{Experimentation}
\subsection{Experimental Settings}
\subsubsection{Datasets.}
Experiments are conducted on three benchmark datasets: MSCOCO~\cite{chen2015microsoft}, Flickr30k~\cite{young2014image}, and SS1M~\cite{feng2019unsupervised}. We follow Karpathy~\cite{karpathy2015deep} to split the MSCOCO and Flickr30k datasets into the training, validation, and test sets. Only text annotations from the training set are used for text-only training. SS1M is a web-crawled corpus, designed for MSCOCO captions. It relies on the names of object classes in MSCOCO as keywords to gather 2,322,628 unique image descriptions from Shutterstock. We use this corpus and additionally filter out sentences with more than fifteen words, resulting in 978,662 descriptions. More detailed experimental settings and results are given in the supplementary material.

\subsubsection{Evaluation Metrics.}
Following common settings~\cite{tewel2022zerocap,li2023decap}, there are several metrics considered to evaluate the generated caption, including BLEU-4 ({\bf B@4})~\cite{papineni2002bleu}, METEOR ({\bf M})~\cite{banerjee2005meteor}, ROUGE-L ({\bf R-L})~\cite{lin2004rouge}, CIDEr ({\bf C})~\cite{Vedantam_2015_CVPR}, and SPICE ({\bf S})~\cite{anderson2016spice}.

\begin{table*}[t]
\centering
\resizebox{1.55\columnwidth}{!}{
\begin{tabular}{l|ccccc|ccccc}
\toprule
\multirow{2.5}{*}{Method}&\multicolumn{5}{c|}{MSCOCO}&\multicolumn{5}{c}{MSCOCO$\rightarrow$Flickr30K}\\
\cmidrule{2-11}
  &\textbf{B@4}($\uparrow$) & \textbf{M}($\uparrow$) & \textbf{R-L }($\uparrow$) & \textbf{C}($\uparrow$) & \textbf{S}($\uparrow$)&\textbf{B@4}($\uparrow$) & \textbf{M}($\uparrow$) & \textbf{R-L }($\uparrow$) & \textbf{C}($\uparrow$) & \textbf{S}($\uparrow$)\\
\midrule
Baseline  &25.1  &23.8 &49.4 &88.5&17.7 &15.9  &17.5 &41.0 &34.8 &11.3\\

+Feature Optimization (FO)  &26.2 &24.0 &50.2 &89.8 &17.8 &16.3 &18.1 &42.0 &35.5 &11.3\\
+Feature Projection (FP) &28.5 &25.8 &52.7 &98.2 &19.3 &16.9 &18.0 &42.2 &36.2 &11.3\\
+Auxiliary Feature (AF) &25.5 &24.2 &49.6 &91.2 &18.2 &16.9 &17.6 &41.6 &34.7 &11.2\\
\midrule
+FO \& AF w/o FP  &26.3 &24.4 &50.8 &92.1 &18.2 &16.5 &17.8 &41.0 &35.4 &11.2\\
+FP \& FO w/o AF  &29.1 &25.7 &52.9 &100.0 &19.4 &17.1 &18.4 &42.5 &36.8 &11.5\\
+FP \& AF w/o FO  &29.1 &\textbf{26.0} &53.1 &100.3 &\textbf{19.6} &16.9 &18.3 &42.4 &37.1 &11.7\\
\midrule
SynTIC   &\textbf{29.9} &25.8 &\textbf{53.2} &\textbf{101.1} &19.3 & \textbf{17.9}  & \textbf{18.6} & \textbf{42.7} &  \textbf{38.4} &\textbf{11.9}\\
\bottomrule
\end{tabular}
}
\caption{Ablation study results on MSCOCO under in-domain and cross-domain settings.}
\label{tab4}
\end{table*}

\subsubsection{Implementations.}
For synthetic image generation, we utilize Stable Diffusion v1-5~\cite{Rombach_2022_CVPR} as the text-to-image model, which leverages 20 sampling steps to generate a 512$\times$512 image for each input text. A pre-trained CLIP VIT-B/32 model is used as the feature extractor. The Adam optimizer~\cite{kingma2014adam} is employed to optimize parameters. For better alignment, we optimize pseudo-image features using Eq.~\ref{eq2} with a temperature $\tau$ of 1/100 and a learning rate of 1e-5. We use a transformer decoder structure with 4 layers and 4 attention heads as the caption generator and train it from scratch. The projection temperatures $\tau$ in Eq.~\ref{eq3} for MSCOCO, Flickr30k, and SS1M are set to 1/100, 1/80, and 1/100, respectively. A pre-trained object detection model, DETR~\cite{carion2020end}, is used to extract the objects in images.

\subsubsection{Baselines.}
Supervised and unsupervised methods are set as baselines in the experiments. The fully supervised methods, such as ClipCap~\cite{mokady2021clipcap} and BUTD~\cite{anderson2018bottom}, are considered for in-domain captioning. The unsupervised methods, containing CLIPRe~\cite{su2022language}, MAGIC~\cite{su2022language}, ZeroCap~\cite{tewel2022zerocap}, CapDec~\cite{nukrai-etal-2022-text}, CLOSE~\cite{gu2023can}, and DeCap~\cite{li2023decap}, are the main comparison baselines crossing different captioning scenes.

\subsection{In-domain Unpaired Image Captioning}
Unpaired captioning experiments are first conducted to evaluate performance. SynTIC is compared with both supervised and unsupervised methods. For supervised methods, BUTD uses object detection networks to extract visual features for caption generation. ClipCap and the method of~\cite{barraco2022unreasonable} utilize CLIP as the visual encoder. Regarding the unsupervised methods, we take CLIPRe, ZeroCap, MAGIC, CapDec, CLOSE, and DeCap into consideration. Moreover, ESPER-Style~\cite{yu2022multimodal}, \citeauthor{feng2019unsupervised}~(\citeyear{feng2019unsupervised}), and Laina et al.~(\citeyear{laina2019towards}) achieve unpaired image captioning by taking images and captions from MSCOCO as unpaired data.

\begin{figure}[t]
\centering
\includegraphics[width=0.94\columnwidth]{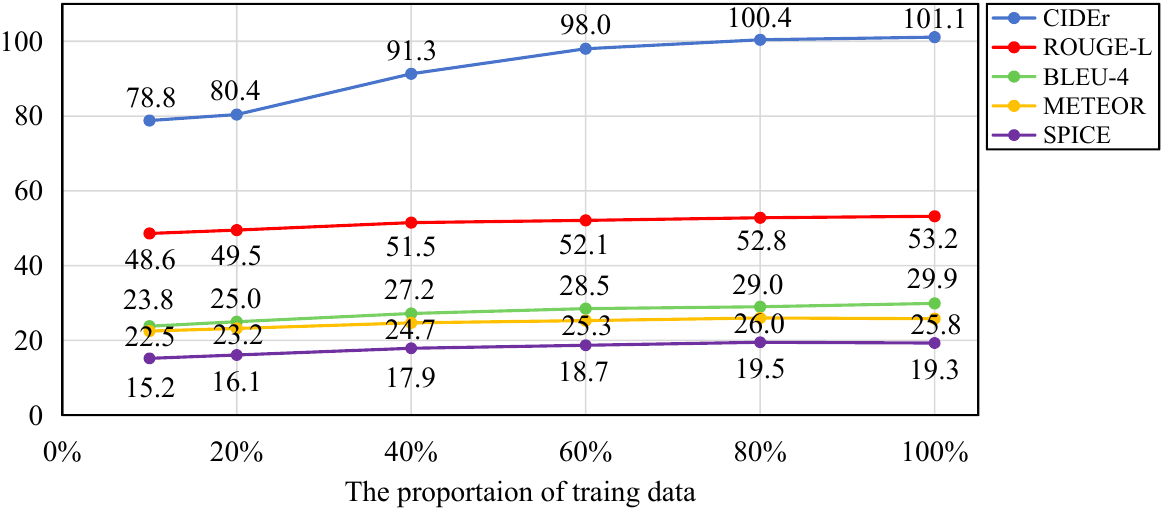}
    \caption{The performance of SynTIC on MSCOCO with different proportions of training data.}
    \label{fig2}
\end{figure}

\subsubsection{Results.}
Table~\ref{tab1} illuminates the in-domain captioning results on the MSCOCO and Flickr30K datasets. Compared with other unsupervised methods, SynTIC exhibits significant performance improvement on most evaluation metrics. showing our method's effectiveness. The unpaired captioning methods could achieve competitive results since they take both images and texts within datasets into consideration. In contrast to CLIPRe, ZeroCap, and MAGIC, text-only methods, including CapDec, CLOSE, DeCap, and SynTIC, obtain better performance by mitigating the modality gap in CLIP. It indicates that the feature gap matters among CLIP-based captioning methods, since this gap degrades performance. Competing with two strong baselines, CLOSE and DeCap, our method increases all metrics on MSCOCO, while also demonstrating performance gains on Flickr30k. The noise injection in CLOSE cannot completely narrow the modality gap. DeCap is hindered by inconsistencies between training and inference, as it only projects image features during inference. Note that SynTIC even outperforms the supervised method on Flickr30k with respect to METEOR and SPICE. It illustrates the strength of bridging multimodal data with synthetic image-text pairs.

\begin{figure*}[t]
\centering
\includegraphics[width=1.8\columnwidth]{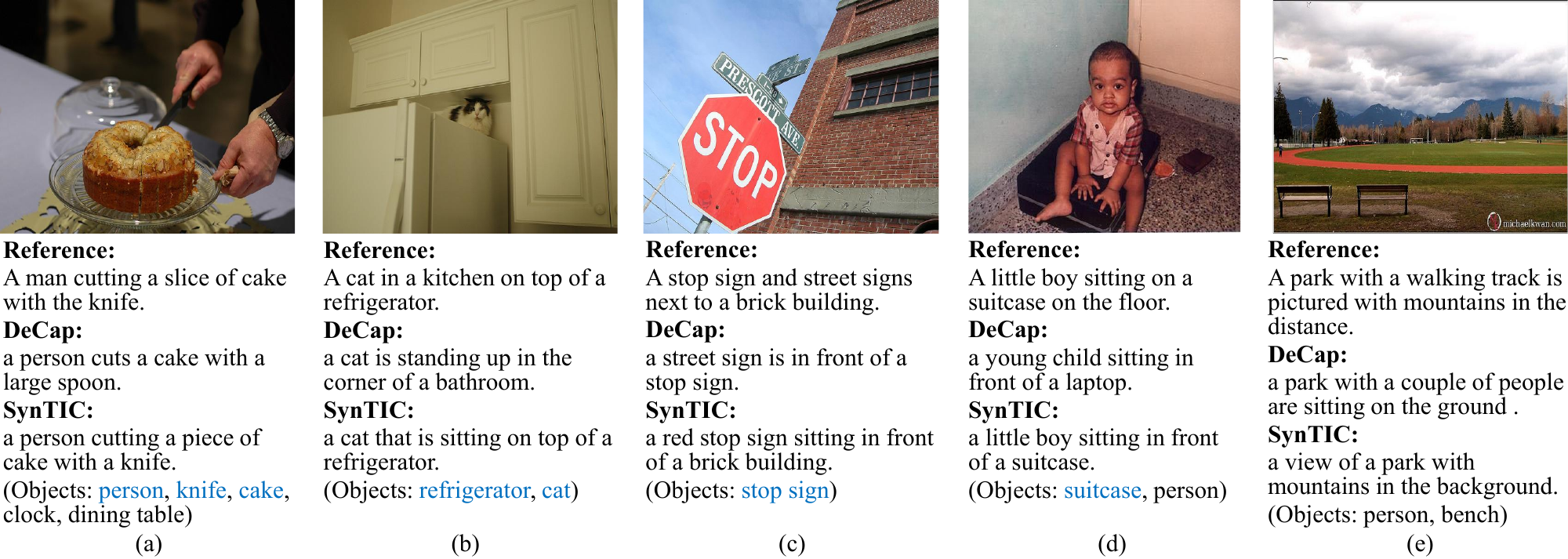}
    \caption{Examples generated from DeCap and SynTIC compared with reference captions on MSCOCO.}
    \label{fig3}
\end{figure*}

\subsection{Cross-domain Image Captioning}
Cross-domain image captioning tests the generalization ability. It requires training the model on a source dataset and directly testing it on a target dataset. SynTIC is compared with various unsupervised methods, including CLIPRe, MAGIC, CapDec, DeCap, and DeCap-TT, where TT represents a feature projection variation using target domain captions.
\subsubsection{Results.}
As shown in Table~\ref{tab2}, our methods consistently outperform baselines on various metrics under both cross-domain settings. Specifically, SynTIC demonstrates significant superiority over CLIPRe and MAGIC. Compared with CapDec, our method beats it in 8 out of 10 metrics, while having similar results on METEOR and ROUGE-L. We attribute CapDec's impressive cross-domain captioning performance to its utilization of the pre-trained language model, GPT2-Base~\cite{radford2019language}. The knowledge derived from GPT2 boosts CapDec through the fine-tuning on text data, especially for the ample captions in MSCOCO. SynTIC deploys a lightweight decoder instead of large language models, which is more computation-friendly while achieving superior results. In contrast to DeCap, SynTIC synthesizes image data to unify training and inference, showing excellent performance. Regardless of the availability of data in the target domain, SynTIC and its variation achieve the best results on all metrics, illustrating that synthetic pairs enhance unsupervised captioning.

\subsection{Zero-shot Image Captioning}
Both in-domain and cross-domain image captioning rely on human-annotated text data with rich prior knowledge and specific descriptive styles (e.g., texts in MSCOCO). Thus, it cannot comprehensively demonstrate the model's zero-shot capability. In this section, we utilize a web-crawled text corpus, SS1M, for training and verify the zero-shot captioning performance on MSCOCO. ZeroCap, CLIPRe, and DeCap are compared with the proposed method.

\subsubsection{Results.}
The zero-shot performance is shown in Table~\ref{tab3}. ZeroCap relies solely on pre-trained models without fine-tuning and obtains the lowest
caption quality. In contrast, CLIPRe shows improvement, as it uses CLIP to retrieve texts from CC3M~\cite{sharma2018conceptual}. Our method outperforms ZeroCap and CLIPRe significantly across all evaluation metrics. Compared with DeCap, which also uses SS1M, SynTIC greatly improves BLEU-4 by 4.4 and CIDEr by 5, while obtaining competitive performance on other metrics. This demonstrates our method could realize better cross-modal alignment with the help of synthetic images and detected objects. The effectiveness of our method is not contingent on the text data with specific descriptive styles. SynTIC maintains strong performance even on web-crawled data.

\subsection{Ablation Study}
\subsubsection{Effect of Components.}
An ablation study is performed on MSCOCO under both in-domain and cross-domain settings, as presented in Table~\ref{tab4}. To evaluate different components within SynTIC, we establish a basic baseline method (i.e., Baseline) by using generated pseudo image features directly as inputs for training while real-world images for inference, without feature optimization (FO), feature projection (FP), and auxiliary features (AF). Due to the difference between synthetic and real images, Baseline is inferior to SynTIC but already approaching DeCap. The method of optimizing pseudo-image features (i.e., +Feature Optimization) brings performance gains over Baseline, indicating that pseudo image features refined by the contrast loss enhance cross-modal alignment. The method of projecting pseudo image features into the CLIP text space (i.e., +Feature Projection) shows great improvements. It demonstrates that gathering text data to represent image features enriches the semantics and effectively bridges the modality gap. Note that this method beats DeCap, which also shows the usefulness of synthetic pairs. When directly attaching auxiliary object features to Baseline (i.e., +Auxiliary Feature), the performance has limited improvements. The reason may be that objects are detected from images, and thus their features are redundant with the image features to some extent. However, when projected features are used as the decoder input (i.e., +FP \& AF w/o FO), the auxiliary features work efficiently and achieve optimal performance in METEOR and SPICE. It indicates that auxiliary features could attach essential information to projected features, benefiting caption generation and cross-modal alignment. Besides, the combination of other components (e.g., +FO \& AF w/o FP and +FP \& FO w/o AF) also contributes to performance gains. SynTIC achieves the best performance by utilizing the above three components, demonstrating the effectiveness of each component under in-domain and cross-domain settings.

\subsubsection{Effect of the Training Data Size.}
To explore the effect of training data size on SynTIC, we sample data with different proportions from MSCOCO for training. The results are shown in Fig.~\ref{fig2}. The performance of SynTIC declines as the training data decreases, but the performance does not drop catastrophically even when there is only 10\% of the training data available. It demonstrates that SynTIC remains competitive even in scenarios with less text data. Note that, regarding all metrics, our method consistently outperforms MAGIC, CapDec, and DeCap with 10\%, 60\%, and 60\% of training data, respectively. In summary, increasing the size of training data enhances SynTIC, and our method still performs well with a sharp decrease of training data.

\subsection{Quantitative Evaluation}
We compare the captions generated by SynTIC with those from DeCap under the in-domain setting on the MSCOCO dataset, as shown in Fig.~\ref{fig3}. The detected objects are listed, and the used ones are highlighted in blue. SynTIC generates more accurate and contextually grounded captions by leveraging synthetic pairs and detected objects. In detail, our captions contain ``knife'' in case (a), ``refrigerator'' in case (b), and ``suitcase'' in case (d), whereas DeCap generates incorrect words like ``spoon'' in case (a), ``bathroom'' in case (b), and ``laptop'' in case (d). The reason may be that DeCap uses only projected features as the decoder input, causing the lack of some image details, while SynTIC achieves better cross-modal alignment. Besides, DeCap's description in case (c) confuses the positional relationship of objects. In case (e), without useful objects, our method obtains correct background information: ``mountains'', while DeCap misses that part and generates nonessential details about ``a couple of people''. This indicates that the auxiliary feature does not restrict our generated descriptions to recognized objects.

\section{Conclusion}
In this paper, we explore synthetic pairs for text-only image captioning with unified training and inference. Considering that the generated images from simple text are distinct from natural images, a generation-then-refinement procedure is developed to adjust image representation. We first obtain pseudo features of the generated images, and optimize them with contrastive constraint, enforcing pseudo features toward the real ones. Then, textual information from the training corpus is assembled to project pseudo features. This process results in a semantically rich image representation in the text embedding space, contributing to improved cross-modal alignment. Moreover, objects detected in images aid the caption decoder in capturing essential content. Experimental results on several benchmark datasets demonstrate that SynTIC outperforms the state-of-the-art methods.

\bibliography{aaai24}



\section*{Supplementary material}

|                  | MSCOCO  | Flickr30K | SS1M    |
| ---------------- | ------- | --------- | ------- |
| Training Texts   | 566,747 | 145,000   | 978,662 |
| Inference Images | 5000    | 1000      | -       |

Table 1: Dataset statistics.

## Appendix A: Experimental Details

### Dataset Settings

The dataset's statistics for text-only captioning are presented in Table 1. In MSCOCO, each image has at least 5 text annotations. According to the Karpathy split, this dataset is divided into a training set of 113,287 images, a validation set of 5,000 images, and a test set of 5,000 images. We use the 566,747 text annotations from the training set for text-only training. Flickr30K contains 5 text annotations for each image. This dataset consists of a training set of 29,000 images, a validation set of 1,014 images, and a test set of 1,000 images. By collecting all textual training annotations, we obtain 145,000 texts for training. The SS1M dataset is conducted from a crawl of 2,322,628 image descriptions extracted from Shutterstock[1], which employs the eighty object classes from MSCOCO as keywords. As in previous works, this dataset is refined by excluding sentences that surpass fifteen words, yielding a training set of 978,662 descriptions.

### Implementations

**Pseudo Image Feature Synthesis.** Our method uses Stable Diffusion v1-5[2] as the pre-trained text-to-image model to generate a 512×512 image for each training text, and sample steps are set to 20. Then, pseudo image features from CLIP are optimized using Eq. 2, where the temperature is set to 1/100. This optimization is performed by the Adam optimizer, with a batch size of 4,096 and a learning rate of 1e-5. The optimized pseudo image features are projected into the text embedding space as decoding features. The procedure of pseudo image feature synthesis is shown in Algorithm 1.

**Auxiliary Feature Extraction.** Our method utilizes a pre-trained detector, DETR-101[3], to extract object tags in the given images, and the detection threshold is set to 0.9. If there is no detected object, the text "none" is used as the tag. The auxiliary features are obtained using a multi-head attention mechanism module with heads of 12 and a hidden size of 768.

**Prefix Decoding.** The language model utilizes a 4-layer and 4-head transformer decoder with a hidden size of 768. Two feed-forward networks are used to convert the dimensions of pseudo-image features and auxiliary features from 512 to 768, respectively. Detailed hyperparameters for the effective training on three datasets are listed in Table 2. During inference, the beam search is employed to generate captions with a beam size of 5. The experiment code in the sup-

---

[1]https://www.shutterstock.com
[2]https://huggingface.co/runwayml/stable-diffusion-v1-5
[3]https://huggingface.co/facebook/detr-resnet-101

---

**Algorithm 1: Pseudo Image Feature Synthesis**

**Require:** Text corpus $D_t = \{t_1, ..., t_n\}$; pre-trained text-to-image model $G$; image encoder $\text{CLIP}_v$; text encoder $\text{CLIP}_t$; number of optimization iterations $M$; number of batch size $b$.

1: **for** $t_i$ in $D_t$ **do** ▷ Generation
2:     Generate the synthetic image $s_i = G(t_i)$
3:     Extract the pseudo image feature $s_i^v = \text{CLIP}_v(s_i)$
4:     Extract the text feature $t_i^t = \text{CLIP}_t(t_i)$
5:     Save the paired features $D_t^p[i] = (s_i^v, t_i^t)$
6: **end for**
7: **for** $m = 1, ..., M$ **do** ▷ Optimization
8:     Sample a batch of $\{(s_k^v, t_k^t)\}_{k=1}^b$ from $D_t^p$
9:     **for** $(s_i^v, t_i^t) \in \{(s_k, t_k)\}_{k=1}^b$ **do**
10:         Optimize $s_i^v$ via Eq. 2
11:     **end for**
12: **end for**
13: **for** $s_i^v$ in $\{s_k^v\}_{k=1}^n$ **do** ▷ Projection
14:     Compute the projected feature $v_i$ via Eq. 3
15:     Save the projected feature $D_p[i] = v_i$
16: **end for**
17: **return** $D_p$

---

| Hyperparameters        | MSCOCO | Flickr30K | SS1M  |
| ---------------------- | ------ | --------- | ----- |
| Training epoch         | 20     | 20        | 10    |
| Batch size             | 128    | 128       | 128   |
| Optimizer              | Adam   | Adam      | Adam  |
| Learning rate          | 5e-6   | 2e-5      | 5e-6  |
| Warmup step            | 1000   | 1000      | 1000  |
| Projection temperature | 1/100  | 1/80      | 1/100 |

Table 2: Hyperparameters for training.

| Time overhead | MSCOCO | Flickr30K | SS1M |
| ------------- | ------ | --------- | ---- |
| Training time | 4.7h   | 1.2h      | 6.6h |

Table 3: Training time for different datasets.

| Method | Training Parameters |
| ------ | ------------------- |
| MAGIC  | 124M                |
| CapDec | 181M                |
| DeCap  | 68M                 |
| SynTIC | 85M                 |

Table 4: Training Parameters for different methods.

plementary materials will be made publicly available for further research.

## Appendix B: Computational Overhead

### Training Time and Parameters

All experiments run on the equipment with four NVIDIA GeForce RTX 3090 GPUs. The computation time using one

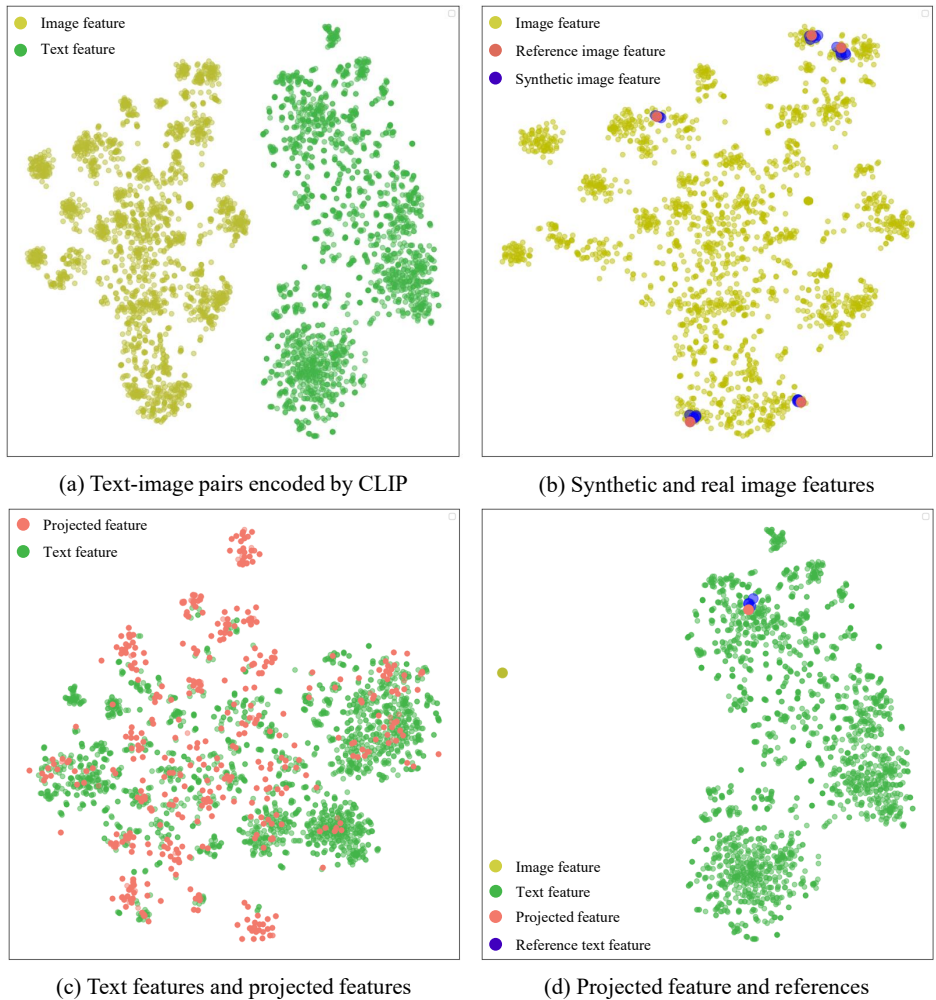

Figure 1: Visualization of features in 2D space using T-SNE.

| Method | Object detection | Image encoding & Caption decoding | All |
|---|---|---|---|
| ZeroCap | - | 5208.9 ms | 5208.9 ms |
| MAGIC | - | 204.5 ms | 204.5 ms |
| CapDec | - | 186.7 ms | 186.7 ms |
| DeCap | - | 44.9 ms | 44.9 ms |
| SynTIC | 61.3 ms | 45.3 ms | 106.6 ms |

Table 5: Inference speed for different methods.

NVIDIA GeForce RTX 3090 for training is listed in Table 3. This shows that the training procedure of our methods is computation-friendly, since the training time on different datasets is invariably less than 7 hours. In addition, we compare the training parameters of SynTIC with recent unsupervised methods: MAGIC, CapDec, and DeCap, as shown in Table 4. By utilizing a lightweight decoder, SynTIC and DeCap have a similar number of training parameters, while our method significantly improves the caption performance. In contrast to MAGIC and CapDec, which fine-tune pre-trained language models for caption decoding, SynTIC has fewer parameters while achieving superior results. This highlights that our method is more efficient.

### Inference Time

The inference speed of different methods is shown in Table 5. Object detection would bring additional overhead, and thus SynTIC has a slower generation speed compared to DeCap. Thanks to the lightweight structure, the inference speed of SynTIC is superior to that of other baselines using pre-trained language models. Our method makes a modest increase in computational cost in exchange for an enhancement in performance. We report the average time cost of captioning 100 images with batch size 1.

## Appendix C: More Experimental Results

### Zero-shot Image Captioning

The performance of SynTIC trained on CC3M is exhibited in Table 6. We construct synthetic pairs using the CC3M text corpus. SynTIC trained on CC3M still beats DeCap trained on CC3M. Moreover, SynTIC trained on SS1M achieves better performance than it trained on CC3M, indicating the

| Method | Dataset | MSCOCO | | | | |
|---|---|---|---|---|---|---|
| | | B@4($\uparrow$) | M($\uparrow$) | R-L ($\uparrow$) | C($\uparrow$) | S($\uparrow$) |
| DeCap | CC3M | 8.8 | 16.0 | - | 42.1 | 10.9 |
| DeCap | SS1M | 8.9 | 17.5 | - | 50.6 | 13.1 |
| SynTIC | CC3M | 11.9 | 16.0 | 35.6 | 46.3 | 11.1 |
| SynTIC | SS1M | 13.3 | 17.6 | 36.7 | 55.6 | 13.3 |

Table 6: Zero-shot image captioning results on MSCOCO.

| Method | Training data | B@4($\uparrow$) | M($\uparrow$) | R-L ($\uparrow$) | C($\uparrow$) | S($\uparrow$) |
|---|---|---|---|---|---|---|
| In-domain Captioning (MSCOCO) | | | | | | |
| SynTIC | Synthetic pairs | 29.9 | 25.8 | 53.2 | 101.1 | 19.3 |
| SynTIC | Real pairs | 32.2 | 27.1 | 54.9 | 107.8 | 20.0 |
| Cross-domain Captioning (MSCOCO$\rightarrow$Flickr30K) | | | | | | |
| SynTIC | Synthetic pairs | 17.9 | 18.6 | 42.7 | 38.4 | 11.9 |
| SynTIC | Real pairs | 16.8 | 17.8 | 41.8 | 36.3 | 11.0 |

Table 7: In-domain and cross-domain captioning results.

web-crawled corpus using keywords from MSCOCO could further improve the performance on associated datasets.

### SynTIC with Real Images

The in-domain and cross-domain captioning performance of SynTIC using real pairs is exhibited in Table 7. For in-domain captioning, the results in the top block show that the fully supervised training enables the model to readily acquire real image-text correspondences, resulting in better performance. The bottom block demonstrates the inferior performance of the model trained on real pairs for cross-domain captioning. SynTIC using synthetic pairs demonstrates a strong cross-domain capability.

## Appendix D: Visualization and Examples

### Feature Visualization

To intuitively observe the text-image alignment, we visualize the text and image features from MSCOCO in 2D space leveraging T-SNE. In Fig. 1(a), there is an obvious modality gap between text features and image features obtained by CLIP encoding. In Fig. 1(b), we highlight several synthetic image features after optimization and their corresponding real image features which are unavailability during training. Note that each real image matches five synthetic images (i.e., five annotated captions). The features of real images are surrounded by related synthetic features, illustrating that alignment learned from synthetic pairs could exhibit real-world image-text association. As shown in Fig. 1(c), our method utilizes the optimization and projection of image features to effectively reduce the modality gap. The overlap between the feature distributions of texts and images demonstrates that cross-modal alignment is improved by SynTIC. We visualize an example in Fig. 1(d), where the projected feature of an image is very close to the text features of human-annotated captions. This also indicates that our method realizes cross-modal alignment efficiently.

### Generated Examples

More examples of generated captions using images from the test set of MSCOCO are shown in Fig. 2. SynTIC is trained on MSCOCO, Flickr30K, and SS1M to obtain captions under the in-domain, cross-domain, and zero-shot settings, respectively.

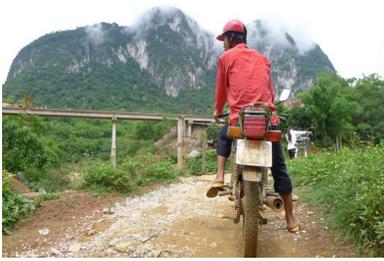 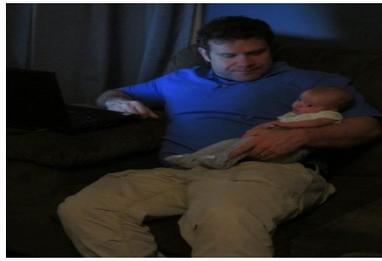 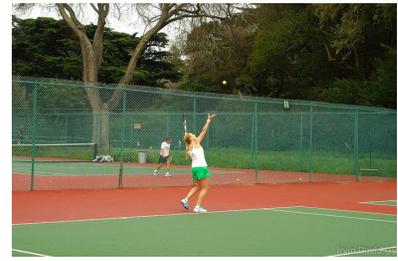

**Reference:**
A man with a red helmet on a small moped on a dirt road.
**In-domain:**
a man is riding a motorcycle down a dirt road.
**Cross-domain:**
a man is riding a motorbike down a dirt road.
**Cross-domain-TT:**
a man on a motorcycle is parked on the side of a road.
**Zero-shot:**
bicycle crossing the river in thailand , vietnam.

**Reference:**
A man uses a laptop computer while holding his baby.
**In-domain:**
a man holding a baby in his lap while holding a laptop.
**Cross-domain:**
a man sitting on a couch holding a baby.
**Cross-domain-TT:**
a man is sitting on a couch holding a baby.
**Zero-shot:**
father with baby using laptop computer at home.

**Reference:**
A woman on a tennis court serves the ball.
**In-domain:**
a woman playing tennis on a tennis court.
**Cross-domain:**
a woman in a white shirt and green shorts is playing tennis.
**Cross-domain-TT:**
a tennis player hitting the ball.
**Zero-shot:**
a female tennis player with a racket on a tennis court.

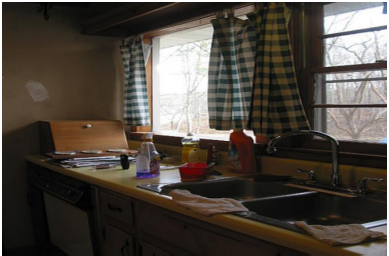 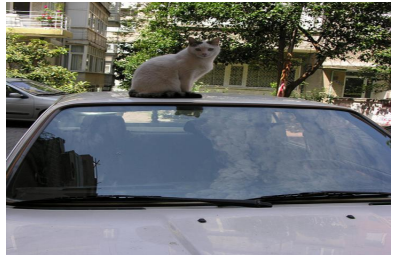 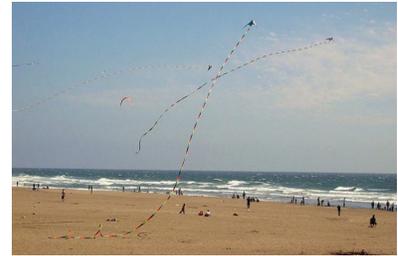

**Reference:**
A kitchen is shown with a variety of items on the counters.
**In-domain:**
a kitchen with a stove , sink and a window.
**Cross-domain:**
a man is cooking in the kitchen.
**Cross-domain-TT:**
an open kitchen with a pot in the kitchen.
**Zero-shot:**
detail of a kitchen with a sink in the kitchen.

**Reference:**
A cat that is on top of a car.
**In-domain:**
a cat sitting on top of a car.
**Cross-domain:**
a black cat sits on a car in the middle of the street.
**Cross-domain-TT:**
a cat is sitting on top of a car.
**Zero-shot:**
the cat is sitting on the roof of the car.

**Reference:**
A group of people on the beach flying some kites.
**In-domain:**
a group of people on the beach flying kites.
**Cross-domain:**
a group of people are flying kite on the beach.
**Cross-domain-TT:**
a group of people are flying kite on the beach.
**Zero-shot:**
silhouette of a kite flying in the sea.

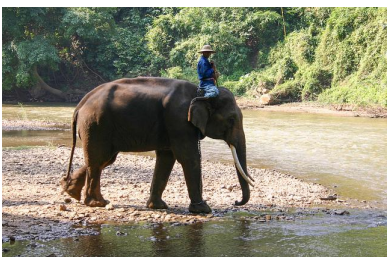 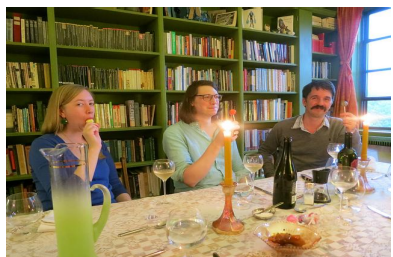 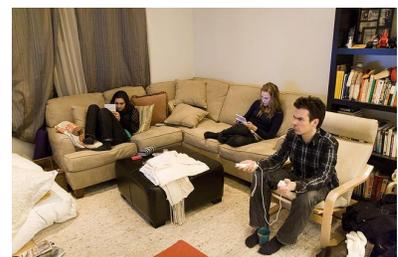

**Reference:**
A man riding a tusked elephant by a shallow river.
**In-domain:**
a man riding an elephant in the middle of a river.
**Cross-domain:**
a man is standing on the side of a river with an elephant.
**Cross-domain-TT:**
a man is standing in front of an elephant.
**Zero-shot:**
elephant on the river in thailand , thailand.

**Reference:**
A group of two women and one man sitting at a table.
**In-domain:**
a group of people at a table with wine glasses.
**Cross-domain:**
a group of people gathered around a table at a table with drinks in front of them.
**Cross-domain-TT:**
a group of people sitting around a table with wineglasses and wine.
**Zero-shot:**
group of friends toasting with glasses of wine at home.

**Reference:**
A group of young people sitting on a couch next to a guy playing a Nintendo Wii.
**In-domain:**
a group of people playing wii in a living room.
**Cross-domain:**
a group of people are sitting in a living room playing a video game.
**Cross-domain-TT:**
a group of people are playing a video game on the couch.
**Zero-shot:**
group of friends sitting on sofa and watching television at home.

Figure 2: Generated captions from SynTIC under different settings on MSCOCO.



\end{document}